\documentclass[letterpaper, 10 pt, conference]{ieeeconf}
\IEEEoverridecommandlockouts
\usepackage{amssymb}
\usepackage{amsmath}
\usepackage{graphicx}
\usepackage{algorithm}
\usepackage{algpseudocode}
\usepackage{cite}
\usepackage{hyperref}
\usepackage{float}
\usepackage{subcaption}
\usepackage{ragged2e}
\usepackage{xcolor}
\newcommand{\revise}[1]{\textcolor{black}{#1}}

\title{\LARGE \bf
 Synergistic Reinforcement and Imitation Learning for Vision-driven Autonomous Flight of UAV Along River
}

\author{Zihan Wang$^{^*}$, Jianwen Li$^{^*}$, and Nina Mahmoudian
\thanks{$^{^*}$Authors contributed equally.}
\thanks{This work was supported by ONR N00014-20-1-2085 }
\thanks{Zihan Wang, Jianwen Li, and Nina Mahmoudian are with the School of Mechanical Engineering, Purdue University, West Lafayette, IN 47907, USA
        {\tt\small wang5044, li3602, ninam@purdue.edu}}%
}

\begin{document}

\maketitle
\thispagestyle{empty}
\pagestyle{empty}

\begin{abstract}
Vision-driven autonomous flight and obstacle avoidance of Unmanned Aerial Vehicles (UAVs) along complex riverine environments for tasks like rescue and surveillance requires a robust control policy, which is yet difficult to obtain due to the shortage of trainable riverine environment simulators.
To easily verify the vision-based navigation controller performance for the river following task before real-world deployment, we developed a trainable photo-realistic dynamics-free riverine simulation environment using Unity.
In this paper, we address the shortcomings that vanilla Reinforcement Learning (RL) algorithm encounters in learning a navigation policy within this partially observable, non-Markovian environment. 
\revise{
We propose a synergistic approach that integrates RL and Imitation Learning (IL). 
Initially, an IL expert is trained on manually collected demonstrations, which then guides the RL policy training process. Concurrently, experiences generated by the RL agent are utilized to re-train the IL expert, enhancing its ability to generalize to unseen data. 
By leveraging the strengths of both RL and IL, this framework achieves a faster convergence rate and higher performance compared to pure RL, pure IL, and RL combined with static IL algorithms. 
 The results validate the efficacy of the proposed method in terms of both task completion and efficiency.}
The code and trainable environments are available\footnote{\href{https://github.com/lijianwen1997/Synergistic-Reinforcement-and-Imitation-Learning}{https://github.com/lijianwen1997/Synergistic-Reinforcement-and-Imitation-Learning}}. 

\end{abstract}

\section{INTRODUCTION}
Autonomous flight of UAVs over inland waterways has been investigated in applications like water sampling \cite{hodgson2022mission}, inspection of irrigation system \cite{ullah2019vision}, and cooperative drone-boat navigation tasks \cite{huang2023usv, li2022robust, gonzalez2021robust}. However, few studies addressed the along-river autonomous navigation tasks of UAVs, which can be applied in rescue, surveillance, and inspection missions over inland waterways. 
Seasonal variations in river channel geometry and water levels render manually planned GPS waypoints less reliable for precise river following missions. Moreover, in cluttered riverine environments, UAVs must navigate while avoiding obstacles such as trees and bridges, all while maintaining a safe altitude above the water.


A camera can be used as the major perception source for the UAV to take advantage of information embedded in images and the recent developments in image processing.
To reduce the cost and safety issues for real-world experiments, we developed a riverine simulation environment using Unity \cite{juliani2018unity} to verify the feasibility of controlling a UAV to fly along a river with only photo-realistic vision input. 

\begin{figure}[t]
    \centering
    \vspace{0.3cm}
    \includegraphics[width=0.35\textwidth]{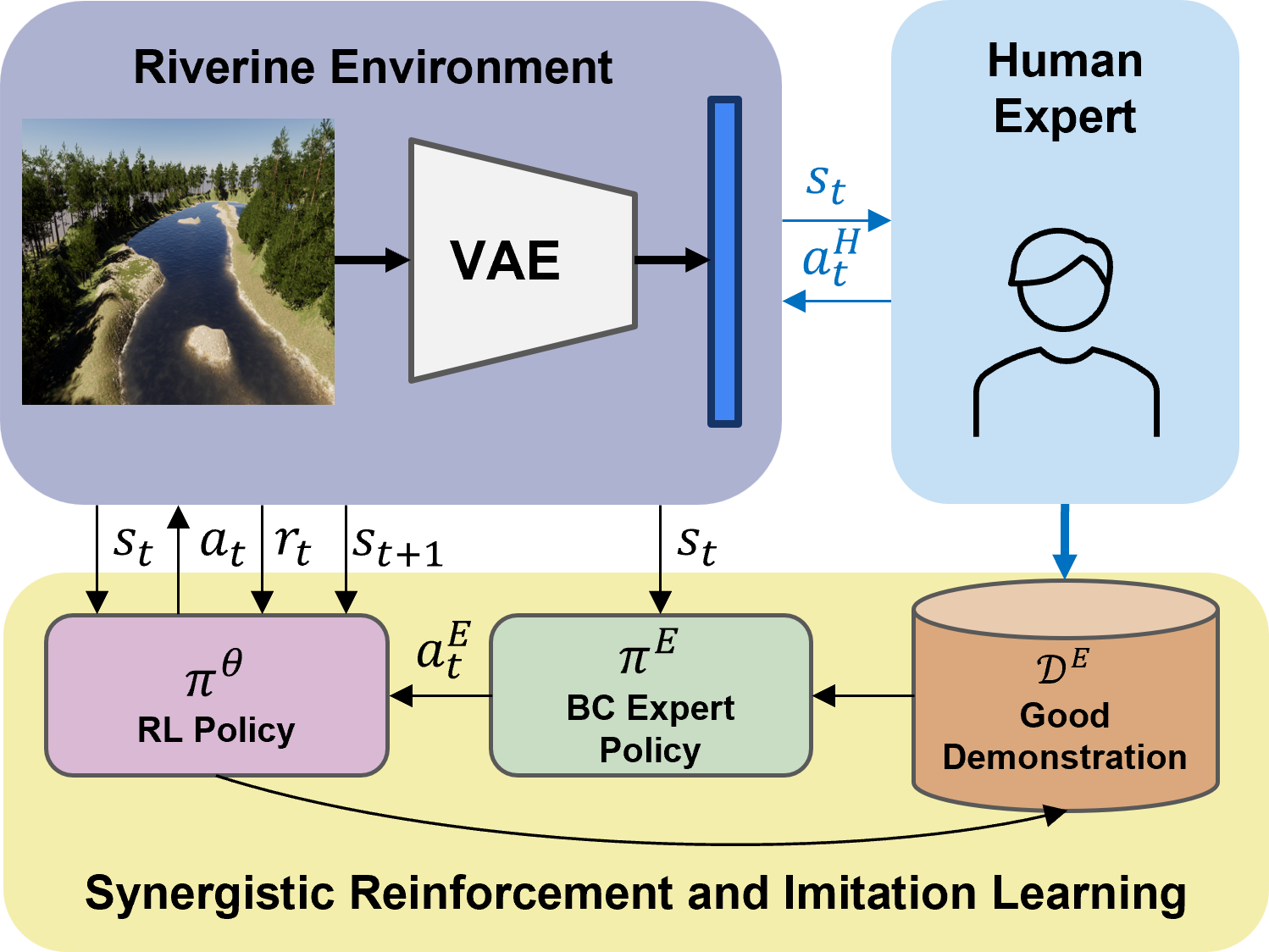}

    \caption{System architecture. A human expert collects good trajectories before training and the transitions are represented by blue arrows. $a^H_t$ denotes the human expert's action. The transitions during training are represented by black arrows. $a^E_t$ denotes the IL expert's action.}
    \label{fig:System}
\end{figure}


\revise{River following can be classified as a track following task, which poses significant training challenges as the environment does not naturally conform to a Markov Decision Process (MDP) due to the agent's partial observability and the non-Markovian reward function \cite{neider2021advice}. Specifically, the first-person camera view of the UAV agent in the riverine scene may lack sufficient water pixels to guide the agent along the river, or it may contain bifurcated river channels, confusing the agent's navigation decisions. 
Moreover, the UAV agent receives positive rewards only upon visiting new river segments, with no reward for idling above already visited segments. 
Consequently, the immediate reward depends on both past observations and actions, rather than solely on the current ones, rendering the task a Partially Observable Markov Decision Process (POMDP) with non-Markovian rewards. 
To address this challenge, combined approaches of RL and Learning from Demonstration (LfD) have been proposed \cite{kang2018policy,jing2020reinforcement,lin2023tag}. However, how to effectively leverage the complementary advantages of imitative learning from demonstrations and reinforcement learning from experience for the track following task remains an area requiring further research and exploration.}

\revise{To investigate into this problem, we pre-train an IL expert using human demonstrations to provide basic guidance, then combine a RL algorithm with the online-updating IL algorithm that accumulates and learns from decent agent experiences, to
mimic the progressive expert behaviors while learning.
In our experiments, Proximal Policy Optimization (PPO \cite{schulman2017proximal}) is chosen as the RL agent, Behavior Cloning (BC \cite{pomerleau1991efficient}) the expert.} 
\revise{This training framework is examined to boost the learning speed and improve final agent performance in the track following tasks, outperforming PPO, StaticBC, DynamicBC and PPO with StaticBC guidance baselines.}


In summary, the main contributions of this paper are:
\revise{
\begin{itemize}
    \item The development of synergistic reinforcement and imitation learning framework that interactively trains RL agent and IL expert to reduce the convergence time and increase the overall performance.
    \item The creation of the trainable photo-realistic riverine simulation environment, which is publicly available, and serves as a testbed for vision-driven autonomous navigation of UAVs in the river following task.
\end{itemize}
}


\section{RELATED WORK}\label{sec:related work}

Autonomous river following in a relatively simple curved river segment has been tested using a water segmentation mask from HSV threshed images \cite{taufik2015multi,taufik2016multi}. 
\revise{This water mask extraction technique and the control method both require careful manual tuning, especially for complex riverine environments with obstacles and multiple river channels. 
To alleviate laborsome manual tuning of the river following controller, modern image embedding methods and RL algorithms can be investigated in a photo-realistic riverine simulator.} 

Game engines as simulation environments have brought more advantages for vision-driven RL training, due to their flexibility in customizing environmental parameters, their low cost as a testbed before real-world deployment, and most importantly, their photo-realism to reduce the Sim2Real gap in air \cite{song2021flightmare, loquercio2021learning}, land \cite{xiang2020sapien, marchesini2020discrete} (indoor/outdoor), and even underwater applications \cite{zielinski20213d}. 
To achieve a safer and more robust autonomous navigation controller in a diverse and adaptive riverine domain, researchers utilized Unreal Engine to train the agent for the river following task \cite{wei2022vision,liang2021vision}. Their proof of the superiority of the Variational Auto Encoder (VAE) \cite{kingma2013auto} for image encoding, over end-to-end controller training, is adopted in our work. 
However, the reset criteria of their river simulation environment are manual-based, the agent action is limited to a 2D plane, and the environment is not publicly available. To overcome these difficulties, our Unity-based riverine simulator can be automatically trained and tested for the river following task, supports 3D action space and collision detection, and is made public as an OpenAI gym \cite{brockman2016openai} environment, thanks to Unity ML-Agents Toolkit \cite{juliani2018unity}.



\revise{RL and IL are potent techniques in robotics. RL enables robots to learn optimal behavior through trial and error, mastering tasks like obstacle avoidance \cite{li2023dynamic}, path planning \cite{aradi2020survey}, and autonomous driving \cite{feng2023dense} even in dynamic environments. Conversely, IL harnesses expert demonstrations to swiftly train robots, allowing them to mimic complex skills. This method is particularly valuable when human expertise is accessible, streamlining robot learning without extensive programming \cite{hussein2018deep}.
However, both approaches have limitations. 
RL training can be time-consuming and costly, often demanding a substantial number of interactions with the environment to learn optimal policies. 
IL's effectiveness is constrained by its reliance on static training datasets, making it challenging to generalize to novel scenarios.} 

\revise{To address these issues, researchers have begun integrating RL and IL to enhance training efficiency and performance. 
For instance, DAgger \cite{ross2011reduction} enhances BC by training on a dynamically updating dataset mirroring the observations the trained policy is expected to encounter. However, it necessitates continual online querying of optimal actions from a human expert.
Other approaches include frameworks that use the guidance of IL to train RL \cite{huang2022efficient, goecks2020integrating}, regulating the discrepancy between the agent's policy and the expert's. 
In particular, \cite{goecks2020integrating} combines BC and 1-step Q-learning losses to smoothly transition from IL to RL without performance degradation. However, for these methods, the IL policy is static, and further improvements of RL agents are impeded by the constraints imposed by the expert policy.} 

\revise{To overcome these limitations, we introduce an interactive framework called "Synergistic Reinforcement and Imitation Learning." This framework trains an RL agent using IL expert advice while simultaneously refining the expert policy through online sampling of high-quality demonstrations generated by the RL agent. By facilitating collaborative improvement between IL and RL methodologies, this approach enhances training efficiency and overall performance.}


\begin{figure}[h]
    \centering
    \vspace{0.3cm}
    \begin{subfigure}[b]{0.45\textwidth}
        \includegraphics[height=3cm,width=0.48\textwidth]{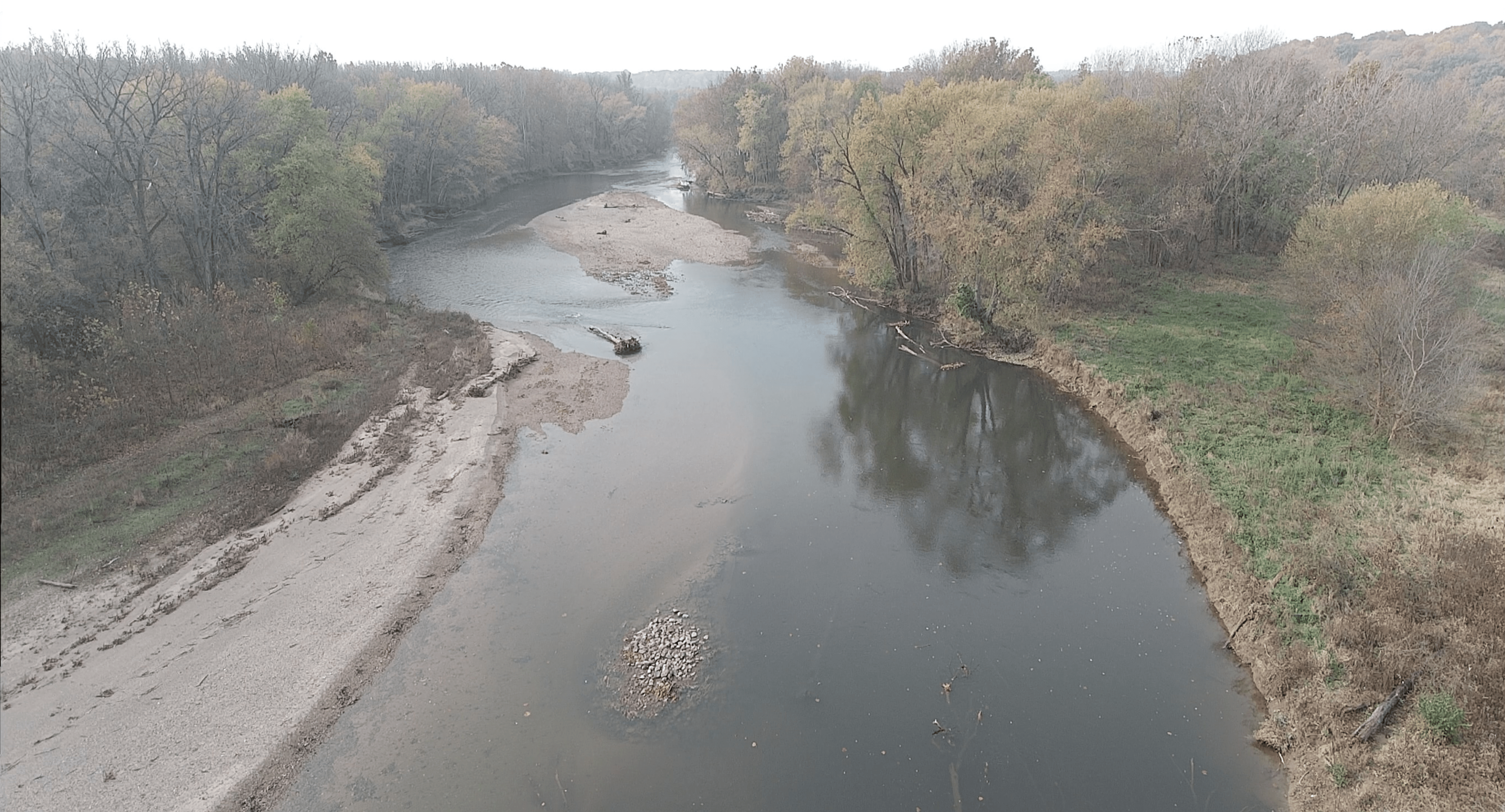} 
        \includegraphics[height=3cm,width=0.48\textwidth]{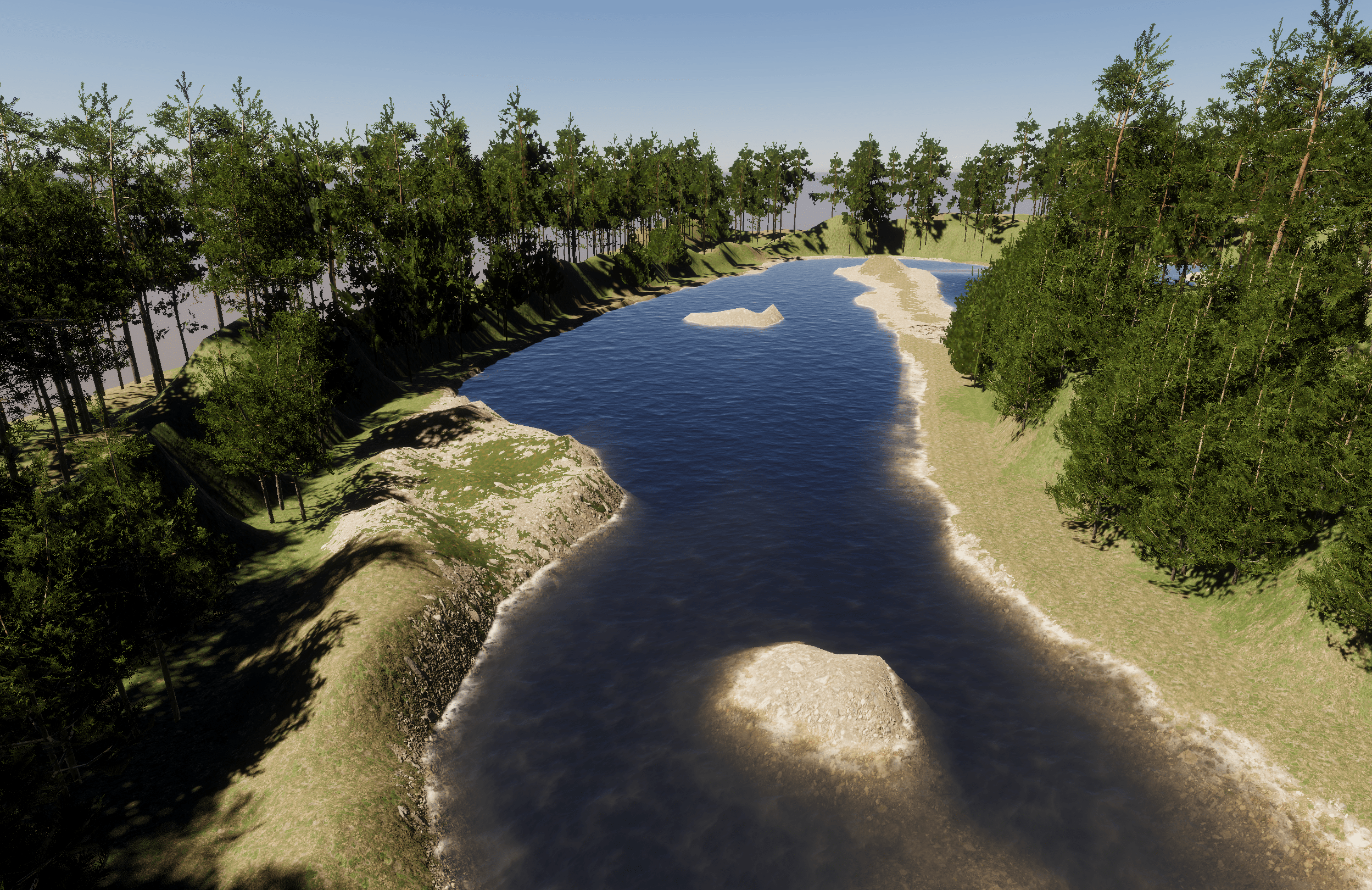}
        \caption{Comparison of an image captured in Wildcat Creek, Indiana, USA (left) and image from Unity river environment (right) that are alike in overall components layout and texture appearance.}
        \label{fig:real_sim_demo_comparison}
    \end{subfigure}
    
    \begin{subfigure}[b]{0.45\textwidth}
        \includegraphics[height=3.2cm,width=0.48\textwidth]{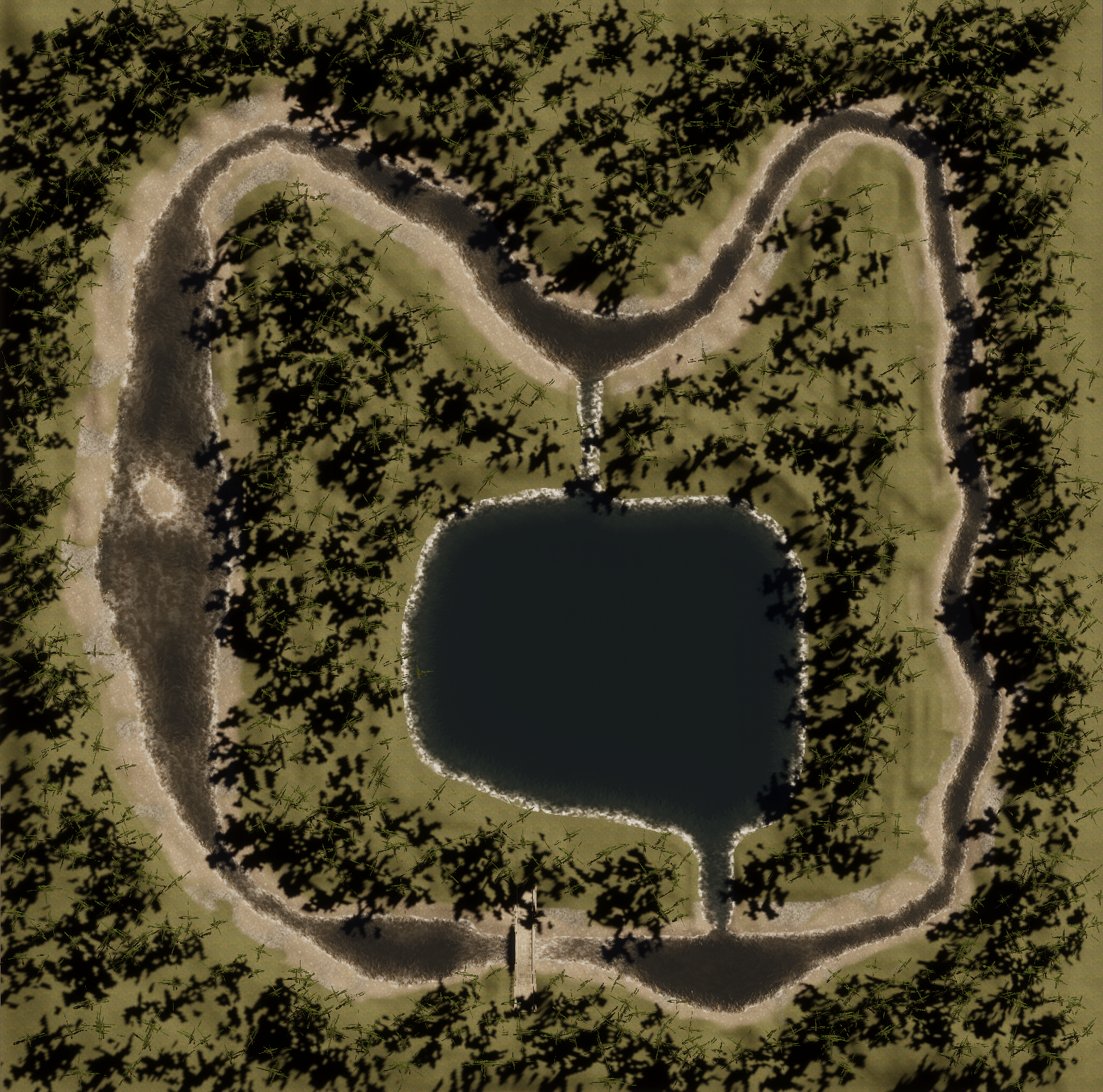}
        \includegraphics[height=3.2cm,width=0.48\textwidth]{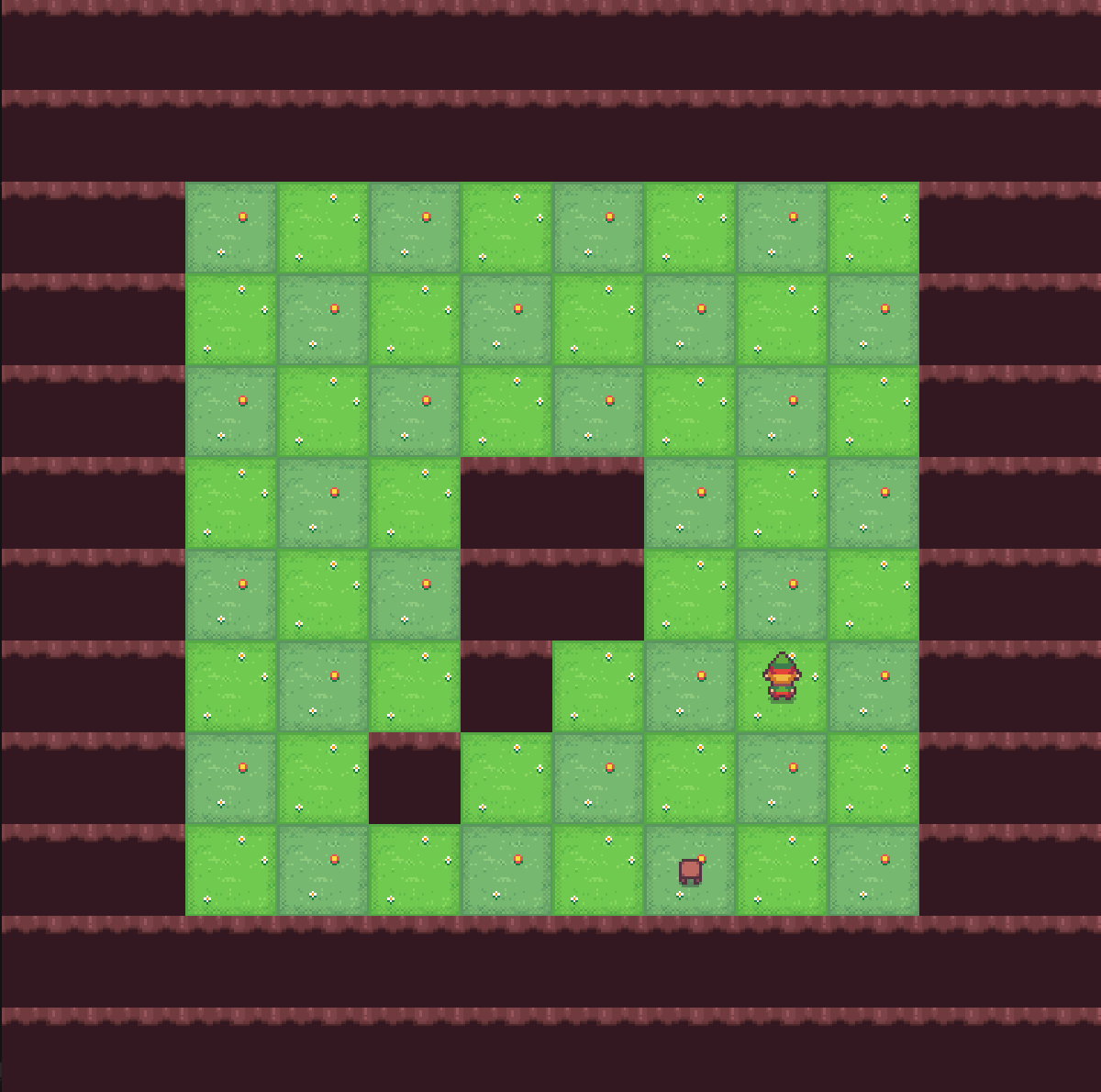}
        \caption{Orthographic images of the riverine simulation environment and the CliffCircular environment.}
        \label{fig:two maps}
    \end{subfigure}
    \caption{Real and simulated riverine environments, and CliffCircular environment (with two extra random cliff grids) for river/track following task.}
\end{figure}

\section{SIMULATION}\label{sec:simulation}
\subsection{Photo-realistic Riverine Environment}
The riverine simulation environment is built in Unity using River Auto Material 2019 
and Mountain Environment 
. The river spline tool is provided by the former package to support fast, flexible, and scriptable river building with various realistic water textures and shaders. The latter package provides a huge library of scanned forest assets like trees, branches, logs, and rocks, which helps to enrich the biomes of the riverine scene, as well as makes it more alike the real-world riverine scene components and appearance, as shown in Fig. \ref{fig:real_sim_demo_comparison}.

Aside from the realistic river and terrain trees, the riverine simulation environment features several additional aspects that not only mimic the general objects in real-world rivers, creeks, and streams, but also make the training of river following agents more difficult: 1) river channels with varied widths and depths, 2) river tributaries that diverge from the main water stream, 3) bridges over the river that can obstruct the navigable path of UAVs, and 4) sharp and slow turns. 

Several environment design choices are made for better photo-realism and more robust training. 
Instead of continuous velocity control, discrete waypoint control for UAV movement is used, which eliminates the efforts of controller dynamics adaptation in real-world riverine environments with various wind speeds, water surface effects, etc.
Thus we will name the UAV agent as the camera agent in the contents hereafter.
The circular shape of the river in our simulated environment is designed for better camera view coherence without seeing too many unrealistic un-modeled parts (sky, terrains, etc.) as the camera agent moves to any valid pose along the river. 
The annulus design of the riverine environment guarantees consistent episodic rewards achievable regardless of the initial reset position of the camera agent at the beginning of each episode.


\begin{table}[h]
    \centering
    \caption{Multi-discrete action values of the camera agent.}
    \begin{tabular}{c|c|c|c}
        Action & 0 & 1 & 2 \\
        \hline \hline
        Up-down Translation (m) & 1 & 0 & -1 \\
        \hline
        Horizontal Rotation (deg) & 10 & 0 & -10 \\
        \hline
        Forward-backward Translation (m) & 1 & 0 & -1 \\
        \hline
        Leftward-rightward Translation (m) & 0.5 & 0 & -0.5 \\
    \end{tabular}
    \label{tab:agent action values}
\end{table}

\subsection{Riverine Camera Agent}
The simulation agent, created using Unity ML-Agents Toolkit \cite{juliani2018unity}, is a camera that looks forward with a fixed pitch angle ($20$ degrees in our experiments) with respect to the drone frame. 

The camera agent is controlled by a four-dimensional multi-discrete action that maps from the command inputs of the common remote controller of UAVs: 1) translation up and down - throttle, 2) horizontal rotation counter-clockwise and clockwise - yaw rate, 3) translation forward and backward - pitch, 4) translation leftward and rightward - roll. 
Each action branch has three discrete values $\{0, 1, 2\}$, where $1$ means no action on that branch. 
The specific values of movement caused by actions are in Table \ref{tab:agent action values}.

\subsection{Riverine Gym Environment}
\begin{figure}[t]
    \centering
    \vspace{0.3cm}
    \includegraphics[width=0.35\textwidth]{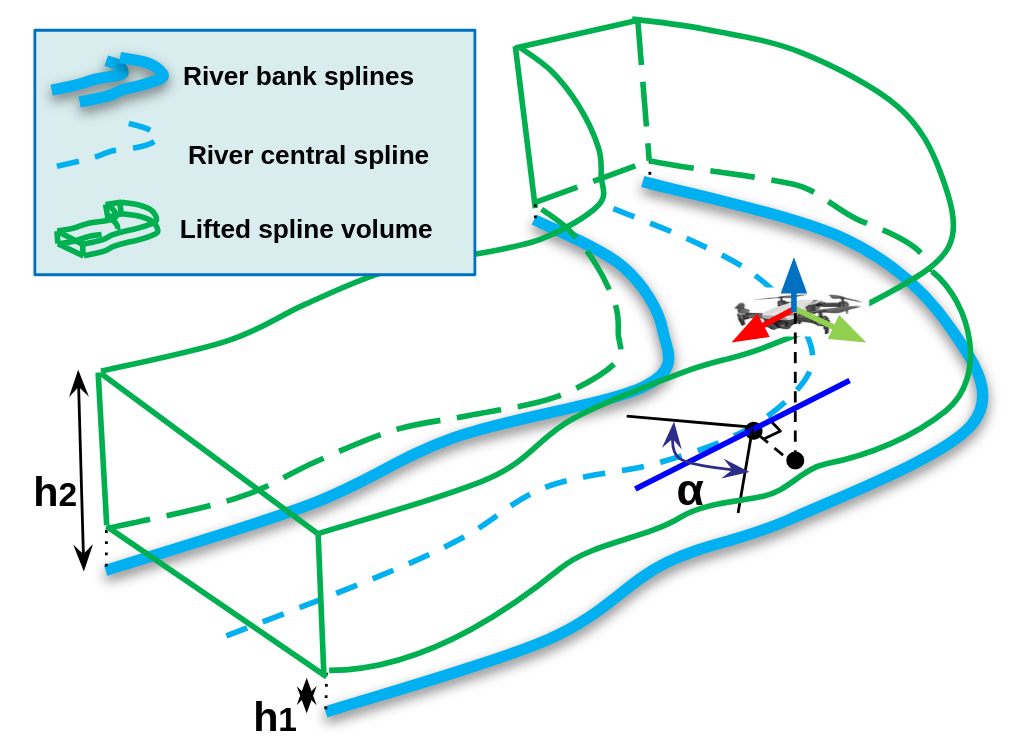}
    \caption[]{Demonstrative diagram of valid activity space and acceptable yaw range of camera agent in the river following task. $h_{1} = 1m, h_{2} = 15m, \alpha = 150^{\circ}$ in our experiments. 
    }
    \label{fig:river spline demo}
\end{figure}

A river can be abstracted as a 3D Catmull-Rom spline \cite{catmull1974class} that is interpolated from several control points. The river surface is formed if we stretch this spline bi-directionally along some direction. 
This closed spline is lifted up to form a 3D bounding volume to limit the height of the camera agent, as well as its horizontal position. 
The tangents along the spline provide failure judgment when the camera agent's yaw angle is far from either tangent angle of the nearest point on this spline with the projected agent position. This prohibits the turnaround motion of the agent since the built river is circular so that the agent can finish the whole-river navigation without any turnaround.
Lingering around or doing continuous meaningless movements with no reward improvement (over $50$ steps in our experiments) will also trigger environment reset, and so does collision with bridges.
If the environment "done" is triggered during training, the camera agent will be reset at a random point inside the spline-based bounding volume, with a random yaw near either spline tangent of the nearest spline point. The above criteria for the river following task can be better visualized in Fig. \ref{fig:river spline demo}.

The river central spline can be simplified as a sequence of connected line segments with nearly the same length. To have a decent baseline RL algorithm to compare, and to intuitively sense the learning capabilities of different learning algorithms, shaped reward is given to the camera agent if the projection of its positions on the spline before and after an action covers any un-visited line segments. The line segment where the camera agent's current position projects onto is denoted as visited. If the camera agent has visited all the river spline segments, it will get the highest reward, then the environment will be reset. For collision detection, a mesh collider is used for bridges, and a box collider with a side length of $0.5$m is used for the camera agent. The overall reward function for the autonomous river following task is given below.


\begin{equation}\label{eqn:reward}
    \begin{aligned}
        r(s, a) = 
        \begin{cases}
        -1, & \text{agent fails} \\
        r_{shaped} & \text{agent covers un-visited spline segment} \\
        0, & \text{otherwise}
    \end{cases}
    \end{aligned} 
\end{equation}
Agent fails when 1) out of spline bounding volume, 2) yaw diverts too much, 3) collides with bridge, and 4) no progress over certain steps.
$r_{shaped}$ is the percentage of newly-visited segments $n$ after the action $a$ over all segments number $N$, multiplied by a constant value, $10$ in this case: $r_{shaped} = 10 * n / N$. 
All negative reward conditions denote the reset of the environment (Eq. \ref{eqn:reward}). 

\revise{
\subsection{CliffCircular Environment}
CliffCircular (Fig. \ref{fig:two maps}) is derived from the CliffWalking environment \cite{towers_gymnasium_2023, sutton2018reinforcement} and tailored for the track following task. 
Like the camera agent in the riverine environment, the agent in CliffCircular must navigate along walkable grids without stepping onto cliffs until it circumnavigates the central cliff block. 
Additionally, CliffCircular presents partial observability and non-Markovian characteristics due to the local observation of cliffs surrounding the agent's current grid, the flexibility in determining the circulation direction, and the nature of the track following task, which necessitates consideration of historical information.}

\begin{figure}[h]
    \centering
    \includegraphics[width=0.28\textwidth]{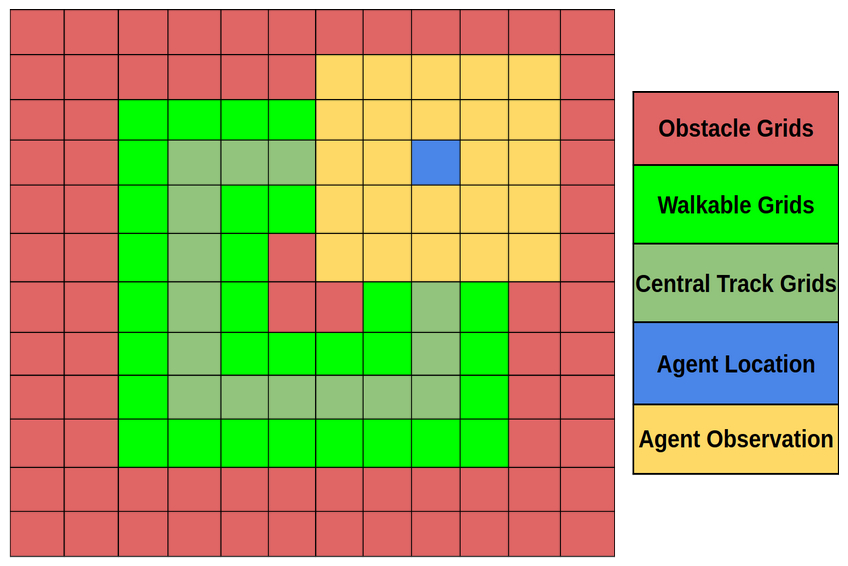}
    \caption{CliffCircular environment diagram. \revise{Central track grids, agent location and observation grids are laid on top of the color layers of obstacle and walkable grids.}}
    \label{fig:circular-diagram}
\end{figure}

\revise{CliffCircular is a $12$x$12$ grid world, illustrated in Fig. \ref{fig:circular-diagram}. The observation space is the flattened $5$x$5$ grids around the agent indicating cliff. Action space includes five discrete operations: no operation, up, right, down and left. The $+1$ reward will be given if agent's projection on the central track grid is newly visited, $-100$ if agent falls off cliff, otherwise $0$. Episode will be reset if cliff is triggered or all central track grids are visited. 
At the beginning of each episode, the agent's location and two additional cliff grids are randomly initialized. This setup encourages the policy to learn a safe and robust track following strategy.
A TimeLimit environment wrapper of length $128$ is used.
}

\section{METHODOLOGY}\label{sec:methodology}

 

The proposed method, illustrated in Fig. \ref{fig:System} and detailed in Algorithm \ref{alg:cap} for solving the river following task, relies on several components: a demonstration dataset provided by a human expert, a BC expert offering guidance, and a PPO agent responsible for environment exploration, dataset updates, and policy learning. The BC expert is firstly trained on the demonstration dataset, $\mathcal{D}^E:\{\tau_i\}_N$ for $N$ successful trajectories\revise{, resulting in the final form of StaticBC and initial form of DynamicBC baselines}. 
Each trajectory is a sequence of state-action pairs $\tau={s_1,a_1,...,s_t,a_t}$. The goal of the expert policy $\pi^E$ is to maximize the log-likelihood over $\mathcal{D}^E$.

\begin{equation}
    \begin{split}
     \pi^E = & {\underset{\pi}{\mathrm{argmax}} \, \underset{(a_t,s_t)\in\mathcal{D}^E}{\mathbb{E}}[\log \pi(a_t|s_t)]]}
    \end{split}
    \label{equ:bc}
\end{equation}

To enhance the performance of the expert policy during PPO training, trajectories with episodic rewards exceeding the reward threshold $T_{reward}$ are added to dataset $\mathcal{D}^E$. 
$T_{reward}$ is set to $4$ in the riverine environment, $18$ in the CiffCircular environment. 
Evaluation of the PPO agent occurs every $10000$ training steps in the CliffCircular environment and every $5000$ steps in the riverine environment, with more frequent updates to the expert policy in the latter due to its higher complexity. In the simpler CliffCircular environment with multi-discrete observation, duplicate transitions are removed from dataset $\mathcal{D}^E$ to mitigate bias. The expert policy undergoes retraining using the latest $2000$ steps from the dataset for $20$ epochs.

\begin{algorithm}
\caption{Synergistic Reinforcement and Imitation Learning}\label{alg:cap}

\begin{algorithmic}

\State Collect the transition dataset $\mathcal{D}^E$ by the human expert
\State Train the IL policy $\pi^E$
\State Initialize policy network $\phi$, value network $\psi$ of RL agent
\Repeat
\State Observe state $s_t$ 
\State Generate action $a_t$ using $\pi^\phi$
\State Execute the action $a_t$ in the environment
\State Observe the next state $s_{t+1}$ and reward $r_t$
\State Store transition $(s_t,a_t,s_{t+1},r_t)$ in the agent buffer $\mathcal{D}^R$
\If{time to update RL agent}
    \State Sample a mini-batch $\mathcal{B}$ from $\mathcal{D}^R$ 
    \State Generate actions $a_t^E$ using $\pi^E$
    \State Calculate the cross entropy $H(\pi^E,\pi^\psi)$ (Eq. \ref{eqn:creosss_entropy})
    \State Calculate loss $\mathcal{L}$ (Eq. \ref{eqn:loss})
   
    \State Update value network $\psi$ and policy network $\phi$

\EndIf
\If{time to evaluate RL agent}
    \State Sample trajectories using $\pi^\phi$
    \If{$\tau$ has a reward above $T_{reward}$ }
        \State Append the trajectory $\tau$ in $\mathcal{D}^E$
         \State Remove repeated transitions from the $\mathcal{D}^E$
        \State Update the IL policy $\pi^E$ (Eq. \ref{equ:bc} )
    \EndIf
\EndIf

\Until Convergence or maximum training steps
\end{algorithmic}
\end{algorithm}
The target of PPO is to obtain a policy, defined as a distribution over actions conditioned on states $\pi(a_t|s_t)$, to maximize the long-term discounted cumulative rewards
\begin{equation}
\underset{\pi}{\mathrm{max}} \, \underset{ \tau\sim p_{\pi}(\tau)}{\mathbb{E}}\Bigl[ \sum_{t=0}^T \gamma^t r(s_t,a_t)\Bigr]
\end{equation}
where $\tau$ is a trajectory and $p_{\pi}(\tau)$ is the distribution of the trajectory under policy $\pi$, and $T$ is the time horizon.
To encourage the PPO policy to be close to the policy learned from the decent demonstrations, we add a penalty term in the loss function
\begin{equation}
\mathcal{L}_{action}=H(\pi^E,\pi^\psi)=-\sum \pi^E(a_t|s_t) \log \pi^\psi(a_t|s_t) 
\label{eqn:creosss_entropy}
\end{equation}
where $H(\pi^E,\pi^\psi)$ is the cross entropy that measures the difference between the expert policy $\pi^E$ and the RL policy $\pi^\psi$.

Combining the above loss functions, \revise{PPO+DynamicBC} becomes
\begin{equation}
\mathcal{L}= w_{1}\mathcal{L}_{policy} + w_{2}\mathcal{L}_{value} + w_{3} \mathcal{L}_{action}
\label{eqn:loss}
\end{equation}
\revise{where we keep the policy loss $\mathcal{L}_{policy}$ and value loss $\mathcal{L}_{value}$ of PPO intact. $w_1$, $w_2$, and $w_3$ represent the weight constants associated with each loss. $w_1$ and $w_2$ are set to 1. $w_3$ is set to $1$ at the beginning. If the PPO achieves a mean score better than BC during evaluation, $w_3$ is set to $0.2$ to encourage exploration. PPO is updated every $1024$ training steps.}
 
\section{SIMULATION VERIFICATION}\label{sec:experiment}

In the riverine environment, observation images are resized to $128$ x $128$ for both encoding and RL training, and the encoded image is a vector of $1024$ float numbers. VAE model is obtained by training $2000$ randomly collected images from the environment. 
The encoded mean vector (without Gaussian noise) is used as the observation input to RL algorithms.
$50$ trajectories each with $50$ steps are collected by a human expert to form the dataset $\mathcal{D}^E$. 
After $50$ epochs of training, the expert policy $\pi^{E}$ has a success rate of $93.27\%$ to predict the expert action from the demonstration and is able to achieve a mean reward of $0.60$ in the riverine environment. This expert policy will be used to generate reference actions during the training of RL agent. 

\revise{In the CliffCircular environment, $100$ trajectories are collected by an expert, which achieves a mean episode reward of approximately $0$. The BC expert, pre-trained on this dataset, attains a reward of $-0.24$, indicating a suboptimal policy that 
is capable of accomplishing the track following objective but commits few errors occasionally.}
Code related to RL and IL is based on StableBaseline3 \cite{stable-baselines3} and Imitation Learning Baseline Implementations \cite{gleave2022imitation}.

\section{RESULTS AND ANALYSIS}\label{sec:results}

\subsection{Training Results}
The proposed method \revise{(PPO+DynamicBC)}, together with baseline methods (\revise{StaticBC, DynamicBC}, PPO, and PPO+StaticBC), are used to train the discrete-action agents in both environments. 
All the methods have been trained for $14000$ steps in the riverine environment and $10000$ steps in the CliffCircular environment for \revise{$5$} different seeds.

\begin{figure}[t]
    \vspace{0.3cm}
    \centering
    \begin{subfigure}[b]{0.47\textwidth}
        \includegraphics[width=0.48\textwidth]{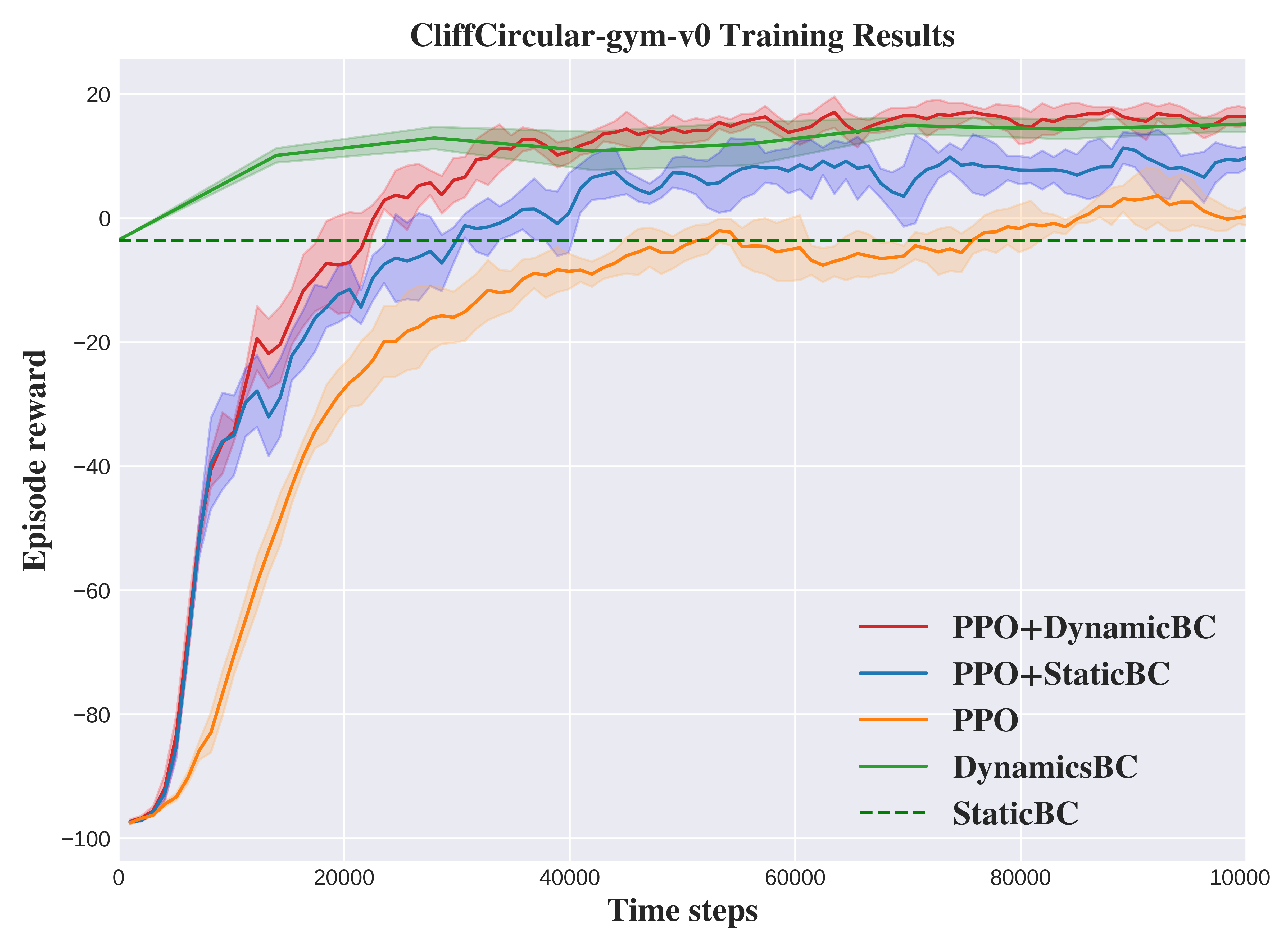}
        \includegraphics[width=0.48\textwidth]{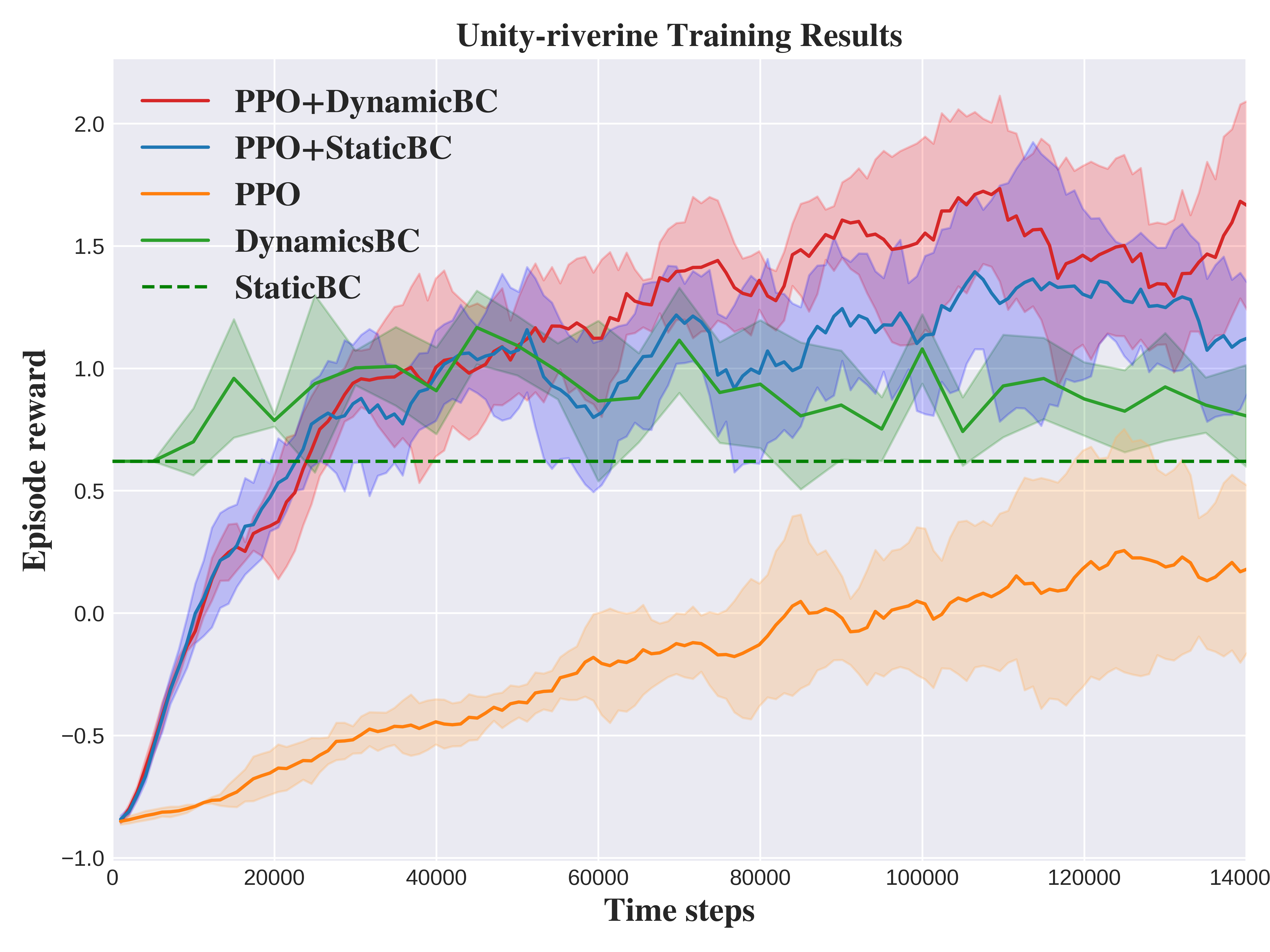}
    \end{subfigure}
  
    \label{fig:training_rewards}

    \caption[]{Training results of different approaches in the CliffCircular environment and river following environment.  }
    \label{fig:training_result}
\end{figure}

Fig. \ref{fig:training_result} illustrates the training results of different methods in two environments. The training curve is derived by averaging results from $5$ different seeds and is depicted as a solid line representing the mean reward, with a shaded region around it indicating the standard error. The episodic reward is computed by averaging over $100$ past episodes, with updates occurring every $1024$ steps. Additionally, the green horizon line signifies the performance achieved by a pre-trained (BC) model.

\revise{In the CliffCircular environment, the PPO baseline slowly improves and surpasses the pre-trained BC baseline after $85000$ steps. The PPO+DynamicBC initially returns rewards similar to PPO+StaticBC baseline. However, PPO+DynamicBC's learning curve shows a higher increasing speed after $10000$ steps and finally reaches a mean episodic reward of $17.90$ at $85000$ steps, while the maximum possible reward is $20$. The best mean episodic reward of PPO+StaticBC is less than $11.31$. After $10000$ steps of training, PPO+DynamicBC's mean episode length is $28.50$, while PPO+StaticBC and PPO's mean episode lengths are $32.62$ and $88.40$, respectively. 
This suggests that the proposed method has acquired a more efficient policy for completing the task in fewer steps, whereas PPO+StaticBC has developed a less optimal policy, requiring an additional $4.12$ steps to complete the task. Evidently, PPO encounters challenges in learning an effective policy within this POMDP track following environment with non-Markovian rewards.
}

\revise{In the river following environment,} PPO baseline slowly improves throughout the reinforcement learning phase but is still lower than the pre-trained BC baseline. The PPO+DynamicBC initially returns rewards similar to PPO+StaticBC baseline. Both methods can get a reward at or above the BC after $25000$ steps. However, after the PPO agent's performance surpasses the BC, PPO+DynamicBC's learning curve shows a higher increasing speed and finally reaches a mean episodic reward of $2.0$ at $105000$ steps, while the best mean episodic reward of PPO+StaticBC is less than $1.5$. PPO+StaticBC also shows multiple decrease trends during training, mainly because of the constraint of the static expert policy. PPO+DynamicBC shows one decrease trend ($110000$ to $13000$) however it recovered to $1.7$ after $10000$ steps. On average, PPO+DynamicBC's score is $1.5$ higher than PPO, $0.6$ higher than BC, and $0.4$ higher than PPO+StaticBC.

\subsection{Evaluation Results}

%



\begin{figure}[t]
    \centering
    \vspace{0.1cm}
    \includegraphics[height=3.5cm,width=0.495\textwidth]{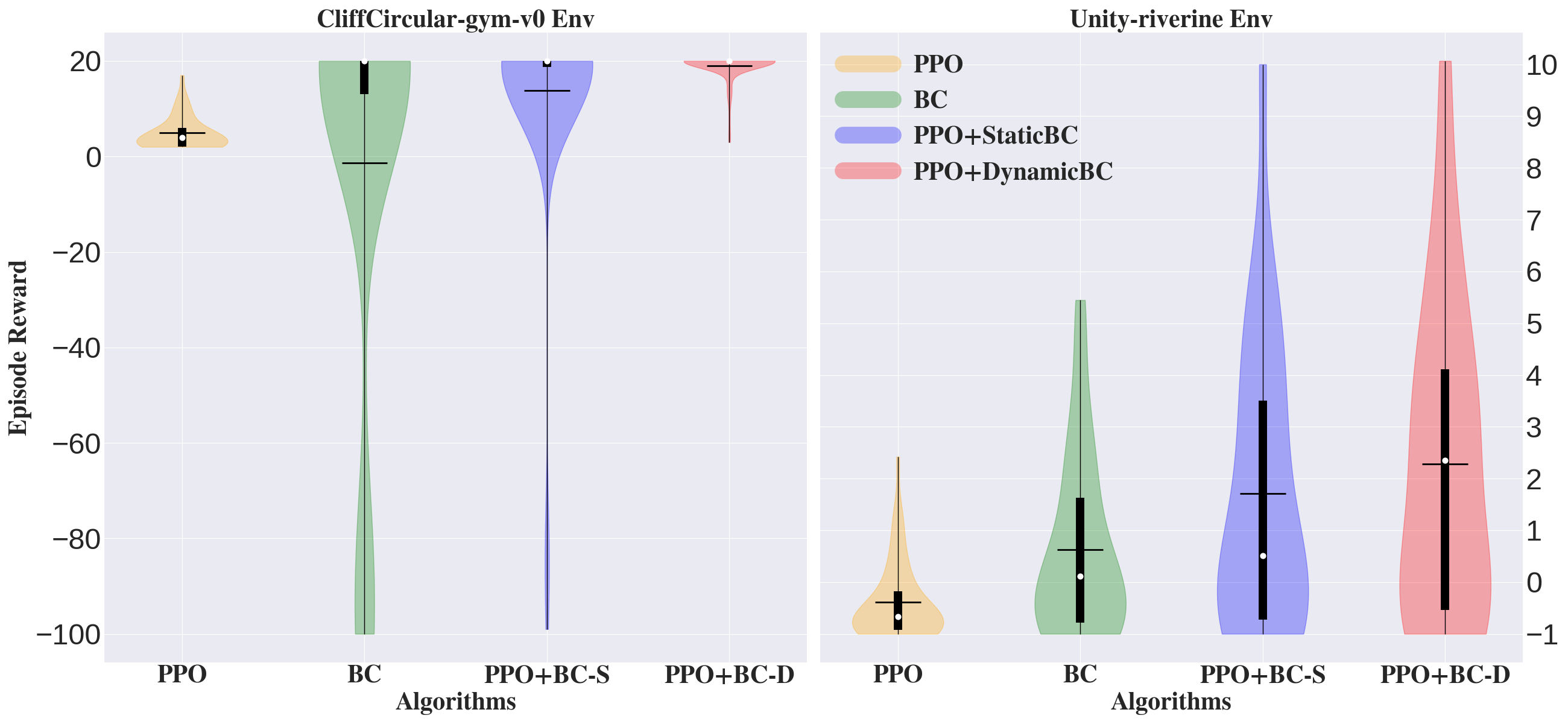}
    \caption{Episode reward distributions of four algorithms on CliffCircular-gym-v0 and riverine environments. The horizontal bar represents the mean, the vertical bar represents 25\% and 75\% percentiles, and the white dot is the median. Statistics from 50 episodes for each algorithm.}
    \label{fig:violin}
\end{figure}

\revise{As shown Fig. \ref{fig:violin}, the reward and navigation efficiency are compared through violin plots. To generate the episodic reward distribution, all methods with the best weights are evaluated for $50$ epochs in the same seed. In the CliffCircular environment, BC has both high and low scores, resulting in a relatively low mean reward. PPO without guidance learns how to avoid collisions but cannot figure out the task objective. PPO+StaticBC is able to increase the chance of finishing the task and decrease the chance of collision compared to StaticBC. PPO+DynamicBC achieves the best result. It learns an optimal policy to finish the track following task while avoiding cliffs. In the riverine environment, BC and the vanilla PPO can never achieve any reward larger than $6$ while both algorithms PPO+StaticBC and PPO+DynamicBC can achieve higher episodic reward of $10$. Noticeably, PPO+DynamicBC achieves the highest mean and median episodic rewards compared to all other methods.}


This verifies the superiority of PPO+DynamicBC algorithm since the IL (BC) model is evolving itself while guiding RL (PPO) model, which acts as a better regularizer compared to PPO+StaticBC algorithm.
Meanwhile, PPO and BC baseline algorithms have inferior performance due to the shortage of runtime expert guidance and limited generalization ability, respectively.
The difference in training convergence and evaluation performance further illustrates the difficulty of the track following tasks. More importantly, this synergistic combination of PPO and BC is validated to largely boost the training speed and final model performance.

\begin{figure}[t]
    \centering
    \vspace{0.25cm}
    \includegraphics[width=0.495\textwidth]{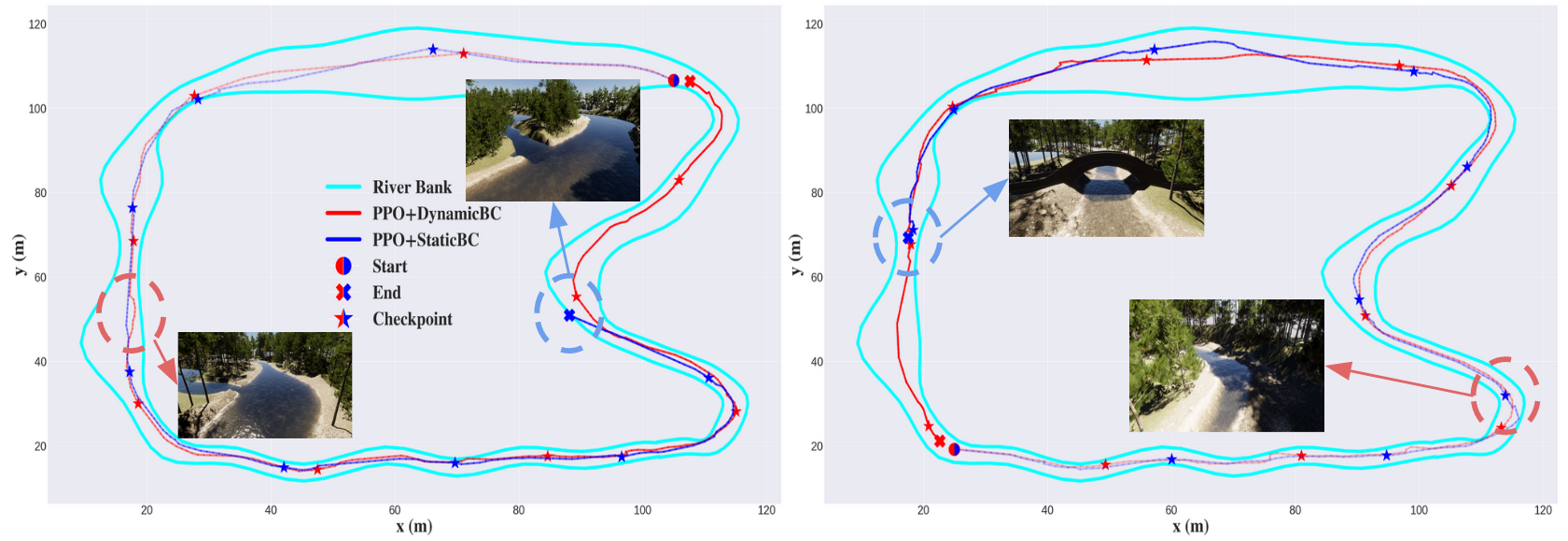}
    \caption{Trajectories of two algorithms PPO+StaticBC and PPO+DynamicBC starting from the same pose in the Unity riverine environment, in two evaluation experiments. Star symbols signify the agent's position every 50 steps. Trajectory color gets darker as the agent moves further away from the starting position.}
    \label{fig:traj}
\end{figure}

To qualitatively compare the navigation efficiency and failure modes of both dynamic and static algorithms, we plot their trajectories in the riverine environment when starting at the same locations (Fig. \ref{fig:traj}). The static algorithm agent can only navigate about $70$\% of river segments in both experiments, before it failed due to exceeding river spline volume (distracted by the tributary \ref{fig:two maps}) and bumping into the bridge, respectively. This is better visualized by the agent view near blue circles.
The dynamic algorithm agent not only finishes whole-river navigation in both experiments but also with fewer steps, which justifies its robustness and efficiency. This can be visualized by the leading checkpoints (star in the figure) of the dynamic algorithm agent, meaning the agent moves a longer effective distance with fewer meaningless movements within the same $50$ steps than the static agent.
Besides, the dynamic algorithm handles better around river corners with smoother trajectory and larger safe margins to river banks. It also rejects tributary when seeing the main stream ahead by timely heading correction. Both cases can be viewed in figures near the red circles in Fig. \ref{fig:traj}. 

\section{CONCLUSION}\label{sec:conclusion}
In this paper, we develop a photo-realistic riverine simulation environment to train a UAV agent to autonomously follow the river while avoiding obstacles. 
A novel framework that synergistically combines RL and IL is proposed, and evaluated in both a riverine environment and a customized grid world track following environment.
The performance of the proposed method (PPO+DynamicBC), compared with the baseline methods (PPO, StaticBC, DynamicBC and PPO+StaticBC) in both environments, achieves the best qualitative and quantitative results. 
In the riverine environment, it achieves at least $0.4$ more rewards than all other methods on average, and the highest best and median rewards.
In the CliffCircular environment, this algorithm finishes the track following task with nearly full score (episode reward $19.04$ out of $20$), with significantly lower reward variance than all baseline algorithms.

\revise{
Further research can improve the proposed method in several aspects.
Firstly, instead of relying solely on the episodic reward threshold to filter trajectories, specific transitions can also be evaluated by some advantage measure and the valuable ones can be merged to the dataset for IL expert refinement. 
This could reduce the demonstration buffer size, bias of this dataset, and the overall training time.  
Secondly, IL expert's policy might not consistently outperform that of the RL agent during training process.
Therefore, when to utilize the action loss for updating the RL agent could be further investigated.
Thirdly, observation augmentation (with historical observations or actions) techniques can be implemented to push the partially observable non-Markovian decision process to MDP for convergence guarantee in the track following tasks.
}


\newpage
\bibliographystyle{IEEEtran}
\bibliography{reference.bib}

\end{document}